\documentclass{vgtc}                          

\ifpdf
  \pdfoutput=1\relax                   
  \usepackage{graphicx}                
  \DeclareGraphicsExtensions{.pdf,.png,.jpg,.jpeg} 
\else
  \usepackage{graphicx}                
  \DeclareGraphicsExtensions{.eps}     
\fi%

\graphicspath{{figures/}{pictures/}{images/}{./}} 

\usepackage{microtype}                 
\PassOptionsToPackage{warn}{textcomp}  
\usepackage{textcomp}                  
\usepackage{mathptmx}                  
\usepackage{times}                     
\usepackage{cite}                      
\usepackage{tabu}                      
\usepackage{booktabs}                  
\usepackage{amsmath}
\usepackage{amssymb}

\onlineid{1129}

\vgtccategory{Research}

\vgtcinsertpkg

\title{A Neural Virtual Anchor Synthesizer based on Seq2Seq and GAN Models}

\author{Zipeng Wang\thanks{e-mail: kohou.wang@cloudminds.com}\\ %
     \scriptsize Clouminds %
\and Zhaoxiang Liu\thanks{corresponding author, e-mail: robin.liu@cloudminds.com}\\ %
     \scriptsize Clouminds %
\and Zezhou Chen\thanks{e-mail: sawyer.chen@cloudminds.com}\\ %
     \scriptsize Clouminds %
\and Huan Hu\thanks{e-mail: hans.hu@cloudminds.com}\\ %
     \scriptsize Clouminds %
\and Shiguo Lian\thanks{e-mail: sg\_lian@163.com}\\ %
     \scriptsize Clouminds %
     }

\abstract{This paper presents a novel framework to generate realistic face video of an anchor, who is reading certain news. This task is also known as Virtual Anchor. Given some paragraphs of words, we first utilize a pretrained Word2Vec model to embed each word into a vector; then we utilize a Seq2Seq-based model to translate these word embeddings into action units and head poses of the target anchor; these action units and head poses will be concatenated with facial landmarks as well as the former $n$ synthesized frames, and the concatenation serves as input of a Pix2PixHD-based model to synthesize  realistic facial images for the virtual anchor. The experimental results demonstrate our framework is feasible for the synthesis of virtual anchor.
} 

\CCScatlist{
  \CCScatTwelve{Computing methodologies}{Computer graphics}{Graphics systems and interfaces}{Mixed/augmented reality};
  \CCScatTwelve{Computing methodologies}{Computer graphics}{Image manipulation}{Image-based rendering}
}

\begin{document}



\maketitle
\begin{figure*}
\begin{center}
   \includegraphics[width=0.8\linewidth]{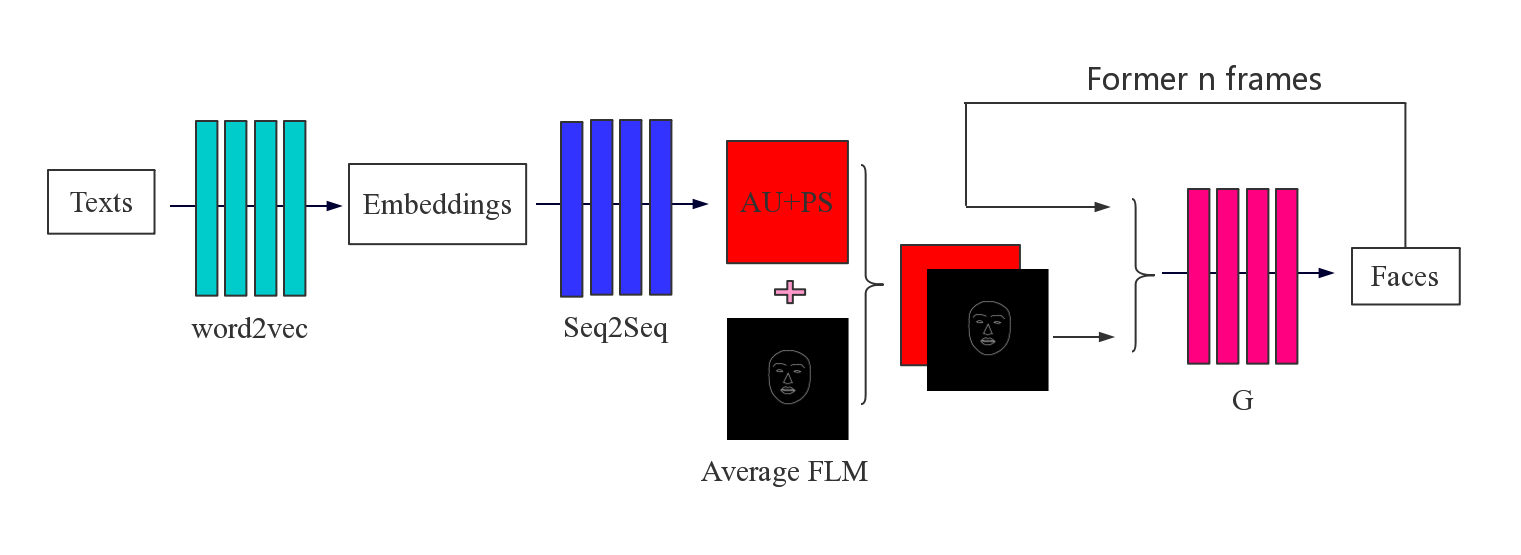}
\end{center}
   \caption{Flowchart of our whole framework. Firstly we utilize Word2Vec to embed words into vectors, then use Seq2Seq to translate vectors into AU+PS. Finally a Pix2PixHD-based generator G will use these generated AU+PS and average FLM as well as the former $n$ synthesized frames to synthesize face images.}
\label{fig3}
\end{figure*}
\section{Introduction} 

Recently, the synthesis of images has received more and more attentions~\cite{isola2017image-to-image, wang2018high-resolution, zhu2017unpaired, wang2018video-to-video, DBLP:journals/corr/abs-1908-06607}, especially after the proposal of Generative Adversarial Network \cite{goodfellow2014generative}. Among which, some methods explored the tasks of translating images of a domain to another~\cite{jin2017AnimeGAN, wang2018pix2pixHD, karras2018styleGAN, palsson2018faceAgingGAN}, while some methods focused on the fidelity of synthesized images~\cite{ledig2017photo, galteri2017deep}. Meanwhile, many other RNN-based methods were also put forward to tackle problems of sequences~\cite{sutskever2014sequence, zhou2015end, krause2016multiplicative, chung2014empirical, mikolov2010recurrent, merity2017regularizing}. However, the task of synthesizing photorealistic face images according to input words, simultaneously ensuring the mouth movements are consistent with the input words, is seldom explored~\cite{matsuyama2016socially, hu2017avatar}.

Unlike previous similar tasks~\cite{hu2017avatar, ichim2015dynamic, saito2017photorealistic,saito2016pinscreen}, we tackle this problem utilizing both GAN-based networks and RNN-based networks, avoiding complicated 3D face model computation. In order to translate input words into corresponding images, we firstly utilize a pretrained Word2Vec model~\cite{mikolov2013efficient} to embed each word into a vector. We then utilize a pretrained Seq2Seq-based model to translate these vectors into corresponding facial action units(AU) and poses(PS) of the target person. These AU and PS, concatenated with facial landmarks and former synthesized frames, are then translated into photorealistic images utilizing a pretrained pix2pixHD-based model~\cite{wang2018high-resolution}. Please note that our work only focuses on visual synthesis, audio synthesis is not considered here.

We take facial AU and PS as an important intermediate representation between input words and synthesized facial images. In training phase, we use a Seq2Seq-based model to train the word embeddings so as to output corresponding AU and PS. For simplicity we will abbreviate AU and PS as AU+PS, and abbreviate facial landmarks as FLM. The ground truth AU and PS are extracted from the person who is speaking those text. In this way, we train the Seq2Seq model to output appropriate AU+PS with words as input. Then for training the pix2pixHD-based model to synthesize the face image of virtual anchor, we concatenate the AU+PS and average FLM and also former $n$ synthesized frames. The average FLM are computed over all the extracted facial landmarks from the training examples. In this way, we utilize the AU+PS to preserve most important information of an individual, meanwhile sidestepping the difficulties of directly translating words into corresponding facial images. Our contributions can be summarized in two aspects:

1.	We propose a novel framework for the synthesis of virtual anchor, whose mouth movements match the corresponding words.

2.	We demonstrate the effectiveness of our proposed framework and the synthesized images' fidelity.

\section{Related works}\label{related}

{\bfseries Face synthesis}.
Garrido et al.\cite{garrido2014automatic} proposed an image-based framework which is conceived as part image retrieval and part face transfer. Their system didn't rely on a 3D face model to map source pose and texture to the target, which excels in simplicity when proposed. Thies~et al.~\cite{thies2016face2face} proposed the first real-time facial reenactment system that requires monocular RGB input only, whose effectiveness was demonstrated in a live setup, where Youtube videos are reenacted in real time.

With the prevalence of Generative Adversarial Network (GAN)~\cite{goodfellow2014generative}, many related methods are proposed to tackle the task of image synthesis, including the synthesis of facial images. Isola~et al. proposed a GAN-based framework to tackle the task of image-to-image translation, which was demonstrated to be effective at synthesizing photos from label maps, reconstructing objects from edge maps, and colorizing images, among other tasks. Afterwards Wang~et al.~\cite{wang2018pix2pixHD} proposed a framework for synthesizing high-resolution photo-realistic images from semantic label maps, which shows a promising prospective for the synthesis of face images from label maps. Zhu~et al.~\cite{zhu2017unpaired} further proposed a framework for learning to translate an image from a source domain $X$ to a target domain $Y$ in the absence of paired examples, which is a significant advancement compared to the difficulty of obtaining paired training data. After the proposal of CycleGAN, many works based on CycleGAN have been proposed for the synthesis of videos of consecutive facial images~\cite{wu2018reenactgan, wang2018video-to-video, bansal2018recycle-gan, jin2017cyclegan, xu2017face}. Among which, Wang~et al.~\cite{wang2018video-to-video} proposed a tracking method which combines a convolutional neural network with a kinematic 3D hand model, which can synthesize anatomically plausible and temporally smooth hand motions, while Wu~et al.~\cite{wu2018reenactgan} present a ReenactGAN, capable of transferring facial movements and expressions from an arbitrary person's monocular video input to a target person's video utilizing a intermediate boundary latent space, which inspire our method greatly. Our method also utilize intermediate facial landmarks for a more photo-realistic and consecutive synthesis of facial images.

{\bfseries Sequence to Sequence}.
 Since Sutskever~et al.~\cite{sutskever2014sequence}proposed the Seq2Seq model, many relevant methods have been proposed to tackle the task of translating a sequence into another. The Seq2Seq model is an encoder-decoder architecture, with a multilatered LSTM~\cite{hochreiter1997long} to embedding the words into vectors while another deep LSTM decodes the target words from these vectors. Gehring~et al.~\cite{gehring2017convolutional} proposed an architecture entirely based on convolutional neural networks, which is equipped with gated linear units~\cite{dauphin2017language}, residual connections~\cite{he2016deep} and attention mechanism~\cite{bahdanau2014neural}. Vaswani~et al.~\cite{vaswani2017attention} proposed another network architecture Transformer, which is based solely on attention mechanisms without recurrence or convolutions. But for our specific task of translating word embeddings into AU+PS, little related work is proposed. In this work, we utilize the Seq2Seq architecture to conduct our translation from word vectors to corresponding AU+PS.

\begin{figure}
\begin{center}
    \includegraphics[width=0.8\linewidth]{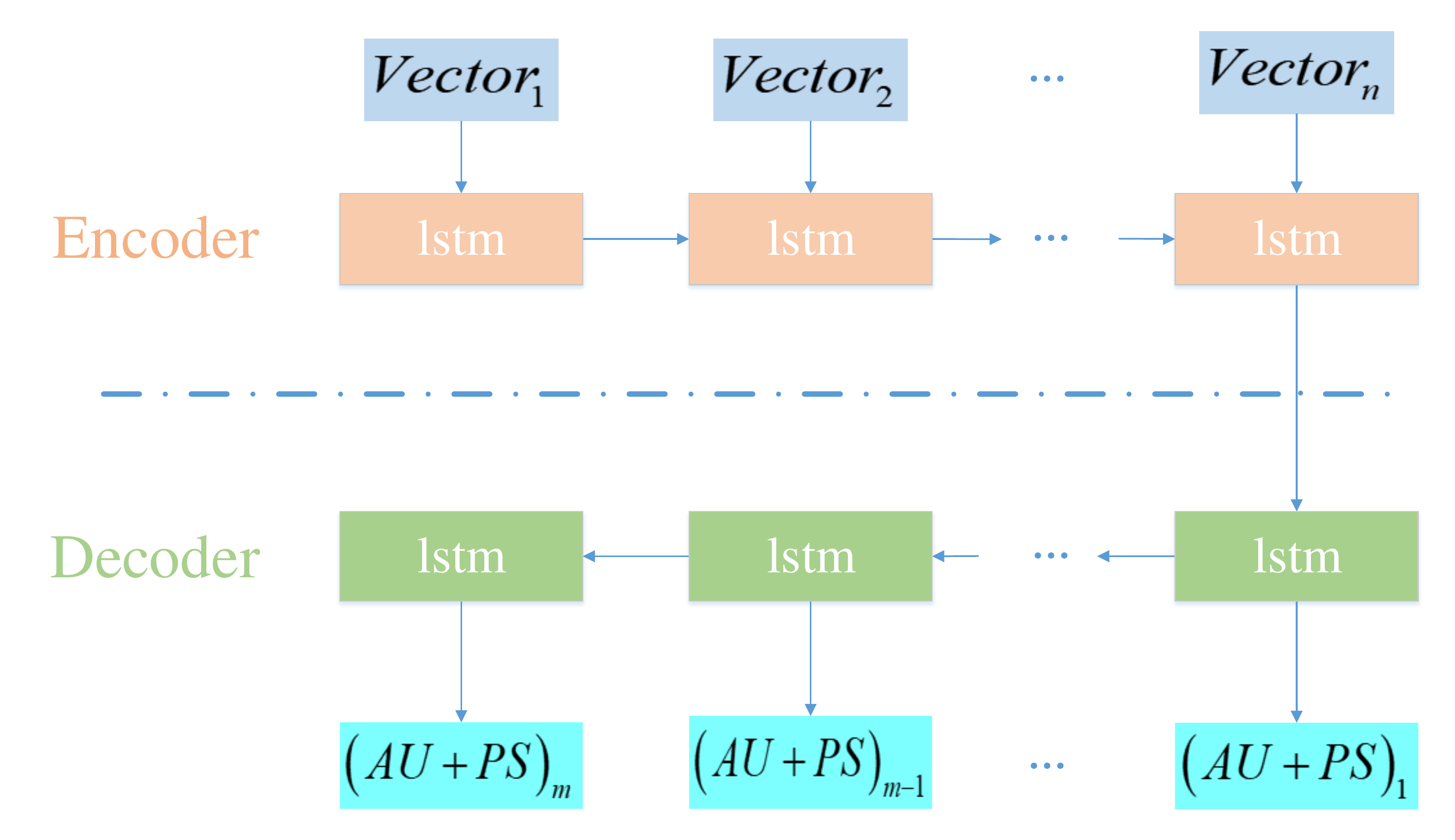}
\end{center}
   \caption{Architecture of our Seq2Seq-based translator. The inputs are word embeddings while the outputs are AU+PS.}
\label{fig1}
\end{figure}

\begin{figure}
\begin{center}
   \includegraphics[width=0.8\linewidth]{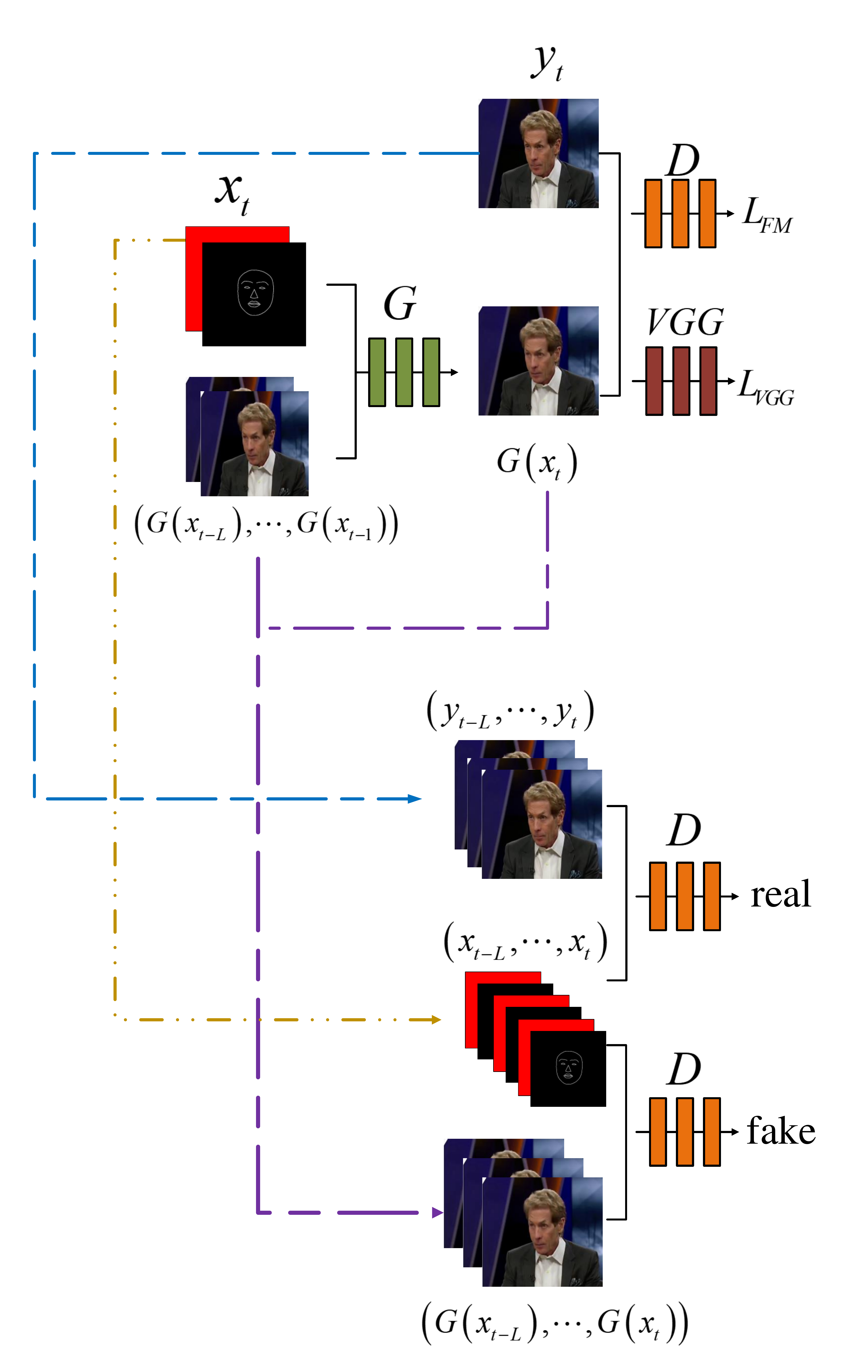}
\end{center}
   \caption{Architecture of our pix2pixHD-based generator. The generator takes AU+PS and average FLM as well as the former $n$ synthesized frames as input and outputs synthesized face image.}
\label{pix2pixHD}
\end{figure}

\section{METHOD OVERVIEW}\label{proposed-methods}

Our method goes like following steps(as shown in Figure~\ref{fig3}): Firstly we embed some texts into vectors using Word2Vec~\cite{mikolov2013efficient}, secondly we utilize a Seq2Seq-based model to translate these word embeddings into AU+PS, then we adopt another pix2pixHD-based network to synthesize face images according to those AU+PS and average FLM as well as the former $n$ synthesized frames. We hold the view that our method can sidestep the difficulty of directly translating words into corresponding facial images by introducing AU+PS and average FLM as intermediate representations. With both of which as a kind of spatial constraints for the training procedure, the GAN-based model can better synthesize more photorealistic and reasonable face images.


The Open-Face~\cite{baltruvsaitis2016openface} provides a convenient way to consistently predict FLM, AU and PS for the face image. Thus, we use it to extract these information from images of target person. The AU+PS is a 20-D vector which comprises of a 17-D AU and a 3-D head pose. On the acquisition of text data, the sentence is obtained directly from the video with python library AutoSub and TTS.

\section{Words to AU+PS}
There exist some popular word embedding methods to represent the text into vectors, including Word2Vec~\cite{mikolov2013efficient}, GloVe~\cite{pennington2014glove}, ELMo~\cite{peters2018deep} and BERT~\cite{devlin2018bert}. Here, for simplicity, we employ a pretrained Word2Vec model to embed each word into a 200-D vector. Then we base our model on the prevalent Seq2Seq~\cite{sutskever2014sequence} architecture to translate these word embeddings into AU+PS(as shown in Figure~\ref{fig1}). In the training phase, the extracted AU+PS of target person is taken as ground truth, while these word embeddings are taken as input.

 Our encoder is defined as
\begin{equation}
\begin{aligned}
    h_{text}^n=LSTM((x_{text}^{n,l})_l)
\end{aligned}
\label{equationx}
\end{equation}
where the LSTM computes the forward sentence encodings, and applies a linear layer on top.

And our decoder is defined as
\begin{equation}
\begin{aligned}
    h_{dec}^l=LSTM(h_{dec}^{l-1}|h_{enc}, y^{l-1})
\end{aligned}
\label{equation0}
\end{equation}
where $h_{dec}$ is the hidden state of $l$-th decoder, and $y^{l-1}$ is the output AU+PS of the $(l-1)$-th decoder. The decoder outputs an AU+PS sequence , which will be used to synthesize the face sequence for the virtual anchor, this procedure will be detailed in Section~\ref{AU+PS2Face}.


\section{Face Synthesis}\label{AU+PS2Face}

 Inspired by \cite{chan2018everybody}, we modify pix2pixHD\cite{wang2018high-resolution} network to generate consecutive face images(as shown in Figure~\ref{pix2pixHD}). We do not directly use AU+PS alone to generate the face sequence. Instead, we combine it with average FLM and the former $n$ synthesized frames to synthesize face sequence.
Compared to AU+PS alone, combining AU+PS with average FLM allows our network to have spatial coordinate constraints during training, which can accelerate the convergence. And the former $n$ frames provide a kind of constraints of temporal correlation of the synthesized frames.


The objective of the generator G is to translate AU+PS maps and average FLM as well as the former $n$ synthesized frames into realistic-looking facial images while the discriminator D aims to determine both the difference in realism and temporal coherence between the "fake" sequence and "real" sequence. The objective function is given by:
\begin{equation}
\begin{aligned}
\begin{split}
    L_{GAN}(G,D)=\mathbb{E}_{(X,Y)}[logD(X,Y)]+\mathbb{E}_{X}[log(1-D(X,G(X)]
\end{split}
\end{aligned}
\label{equation1}
\end{equation}
Where $X$ means the former $n$ (AU+PS,avergae FLM) plus the current (AU+PS,avergae FLM), while $Y$ is the former $n$ ground truth images plus the current ground truth image. Finally, the following combined loss is employed in our task:
\begin{equation}
\begin{aligned}
\begin{split}
    \underset{G}{min}( \underset{D}{max} {L}_{GAN}(G,D)+\lambda_1 *{L}_{FM}(G,D)+\lambda_2*{L}_{VGG}(G(x),y))
\end{split}
\end{aligned}
\label{equation2}
\end{equation}
where, the first term is the adversarial loss in Eq~\ref{equation1}, the second term is feature matching loss~\cite{wang2018high-resolution}, the third term is perceptual reconstruction
loss~\cite{wang2018high-resolution}, and $\lambda_1,\lambda_2$ controls the importance of the three terms.


\section{Experiments}\label{Experiments}


To train our system, we collect more than 400 videos of ESPN shows (First Take and Undisputed) from youtube, these videos all have rich poses and facial expressions which serve as a good ground truth when training.

\begin{figure}
\begin{center}
    \includegraphics[width=0.8\linewidth]{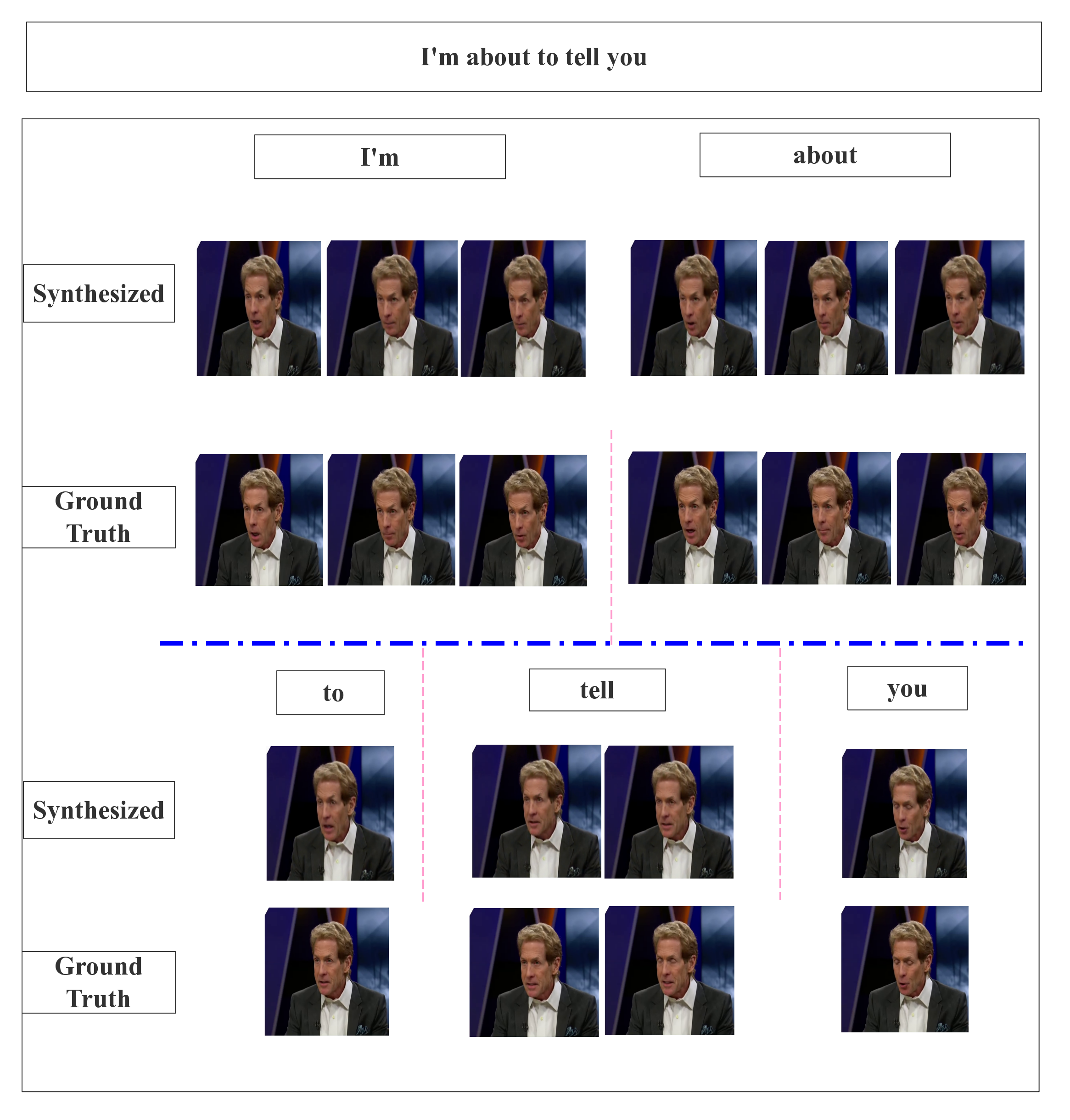}
\end{center}
   \caption{Experiment results. The upper row is the input texts, while the lower row is the synthesized face images speaking corresponding words.}
\label{results}
\end{figure}

In our experiment, we set $n$ as 2. An example result is shown in Figure~\ref{results}. We can see that the synthesized face images are realistic and the mouth movements are almost consistent with the corresponding words. This demonstrates our method's effectiveness.

\section{Conclusion}
We present a novel framework to synthesize consecutive photorealistic video frames according to certain input texts, the synthesized virtual individual has appropriate mouth movements and facial expressions while speaking these input texts, which is known as a Virtual Anchor. The results demonstrate our method's effectiveness and the synthesized video frames are consistent.

{\bfseries Limitations and Future Work}
However, our method suffers from the  problem of efficiency. Due to the frameworks of Seq2Seq and Pix2PixHD, our method is time-consuming and needs a lot of computation resource. Our future work will focus on exploring networks with more lightweight backbones.
\bibliographystyle{abbrv-doi}

\bibliography{template}

\begin{thebibliography}{10}

\bibitem{bahdanau2014neural}
D.~Bahdanau, K.~Cho, and Y.~Bengio.
\newblock Neural machine translation by jointly learning to align and
  translate.
\newblock {\em arXiv preprint arXiv:1409.0473}, 2014.

\bibitem{baltruvsaitis2016openface}
T.~Baltru{\v{s}}aitis, P.~Robinson, and L.-P. Morency.
\newblock Openface: an open source facial behavior analysis toolkit.
\newblock In {\em 2016 IEEE Winter Conference on Applications of Computer
  Vision (WACV)}, pp. 1--10. IEEE, 2016.

\bibitem{bansal2018recycle-gan}
A.~Bansal, S.~Ma, D.~Ramanan, and Y.~Sheikh.
\newblock Recycle-gan: Unsupervised video retargeting.
\newblock {\em european conference on computer vision}, pp. 122--138, 2018.

\bibitem{chan2018everybody}
C.~Chan, S.~Ginosar, T.~Zhou, and A.~A. Efros.
\newblock Everybody dance now.
\newblock {\em arXiv: Graphics}, 2018.

\bibitem{chung2014empirical}
J.~Chung, C.~Gulcehre, K.~Cho, and Y.~Bengio.
\newblock Empirical evaluation of gated recurrent neural networks on sequence
  modeling.
\newblock {\em arXiv preprint arXiv:1412.3555}, 2014.

\bibitem{dauphin2017language}
Y.~N. Dauphin, A.~Fan, M.~Auli, and D.~Grangier.
\newblock Language modeling with gated convolutional networks.
\newblock In {\em Proceedings of the 34th International Conference on Machine
  Learning-Volume 70}, pp. 933--941. JMLR. org, 2017.

\bibitem{devlin2018bert}
J.~Devlin, M.-W. Chang, K.~Lee, and K.~Toutanova.
\newblock Bert: Pre-training of deep bidirectional transformers for language
  understanding.
\newblock {\em arXiv preprint arXiv:1810.04805}, 2018.

\bibitem{galteri2017deep}
L.~Galteri, L.~Seidenari, M.~Bertini, and A.~Del~Bimbo.
\newblock Deep generative adversarial compression artifact removal.
\newblock In {\em Proceedings of the IEEE International Conference on Computer
  Vision}, pp. 4826--4835, 2017.

\bibitem{garrido2014automatic}
P.~Garrido, L.~Valgaerts, O.~Rehmsen, T.~Thormahlen, P.~Perez, and C.~Theobalt.
\newblock Automatic face reenactment.
\newblock In {\em Proceedings of the IEEE Conference on Computer Vision and
  Pattern Recognition}, pp. 4217--4224, 2014.

\bibitem{gehring2017convolutional}
J.~Gehring, M.~Auli, D.~Grangier, D.~Yarats, and Y.~N. Dauphin.
\newblock Convolutional sequence to sequence learning.
\newblock In {\em Proceedings of the 34th International Conference on Machine
  Learning-Volume 70}, pp. 1243--1252. JMLR. org, 2017.

\bibitem{goodfellow2014generative}
I.~Goodfellow, J.~Pouget-Abadie, M.~Mirza, B.~Xu, D.~Warde-Farley, S.~Ozair,
  A.~Courville, and Y.~Bengio.
\newblock Generative adversarial nets.
\newblock In {\em Advances in neural information processing systems}, pp.
  2672--2680, 2014.

\bibitem{he2016deep}
K.~He, X.~Zhang, S.~Ren, and J.~Sun.
\newblock Deep residual learning for image recognition.
\newblock In {\em Proceedings of the IEEE conference on computer vision and
  pattern recognition}, pp. 770--778, 2016.

\bibitem{hochreiter1997long}
S.~Hochreiter and J.~Schmidhuber.
\newblock Long short-term memory.
\newblock {\em Neural computation}, 9(8):1735--1780, 1997.

\bibitem{hu2017avatar}
L.~Hu, S.~Saito, L.~Wei, K.~Nagano, J.~Seo, J.~Fursund, I.~Sadeghi, C.~Sun,
  Y.-C. Chen, and H.~Li.
\newblock Avatar digitization from a single image for real-time rendering.
\newblock {\em ACM Transactions on Graphics (TOG)}, 36(6):195, 2017.

\bibitem{ichim2015dynamic}
A.~E. Ichim, S.~Bouaziz, and M.~Pauly.
\newblock Dynamic 3d avatar creation from hand-held video input.
\newblock {\em ACM Transactions on Graphics (ToG)}, 34(4):45, 2015.

\bibitem{isola2017image-to-image}
P.~Isola, J.~Zhu, T.~Zhou, and A.~A. Efros.
\newblock Image-to-image translation with conditional adversarial networks.
\newblock {\em computer vision and pattern recognition}, pp. 5967--5976, 2017.

\bibitem{jin2017cyclegan}
X.~Jin, Y.~Qi, and S.~Wu.
\newblock Cyclegan face-off.
\newblock {\em arXiv preprint arXiv:1712.03451}, 2017.

\bibitem{jin2017AnimeGAN}
Y.~Jin, J.~Zhang, M.~Li, Y.~Tian, H.~Zhu, and Z.~Fang.
\newblock Towards the automatic anime characters creation with generative
  adversarial networks.
\newblock {\em arXiv preprint arXiv:1708.05509}, 2017.

\bibitem{karras2018styleGAN}
T.~Karras, S.~Laine, and T.~Aila.
\newblock A style-based generator architecture for generative adversarial
  networks.
\newblock {\em arXiv preprint arXiv:1812.04948}, 2018.

\bibitem{krause2016multiplicative}
B.~Krause, L.~Lu, I.~Murray, and S.~Renals.
\newblock Multiplicative lstm for sequence modelling.
\newblock {\em arXiv preprint arXiv:1609.07959}, 2016.

\bibitem{ledig2017photo}
C.~Ledig, L.~Theis, F.~Husz{\'a}r, J.~Caballero, A.~Cunningham, A.~Acosta,
  A.~Aitken, A.~Tejani, J.~Totz, Z.~Wang, et~al.
\newblock Photo-realistic single image super-resolution using a generative
  adversarial network.
\newblock In {\em Proceedings of the IEEE conference on computer vision and
  pattern recognition}, pp. 4681--4690, 2017.

\bibitem{DBLP:journals/corr/abs-1908-06607}
Z.~Liu, H.~Hu, Z.~Wang, K.~Wang, J.~Bai, and S.~Lian.
\newblock Video synthesis of human upper body with realistic face.
\newblock {\em arXiv preprint arXiv:1908.06607}, 2019.

\bibitem{matsuyama2016socially}
Y.~Matsuyama, A.~Bhardwaj, R.~Zhao, O.~Romeo, S.~Akoju, and J.~Cassell.
\newblock Socially-aware animated intelligent personal assistant agent.
\newblock In {\em Proceedings of the 17th Annual Meeting of the Special
  Interest Group on Discourse and Dialogue}, pp. 224--227, 2016.

\bibitem{merity2017regularizing}
S.~Merity, N.~S. Keskar, and R.~Socher.
\newblock Regularizing and optimizing lstm language models.
\newblock {\em arXiv preprint arXiv:1708.02182}, 2017.

\bibitem{mikolov2013efficient}
T.~Mikolov, K.~Chen, G.~Corrado, and J.~Dean.
\newblock Efficient estimation of word representations in vector space.
\newblock {\em arXiv preprint arXiv:1301.3781}, 2013.

\bibitem{mikolov2010recurrent}
T.~Mikolov, M.~Karafi{\'a}t, L.~Burget, J.~{\v{C}}ernock{\`y}, and
  S.~Khudanpur.
\newblock Recurrent neural network based language model.
\newblock In {\em Eleventh annual conference of the international speech
  communication association}, 2010.

\bibitem{palsson2018faceAgingGAN}
S.~Palsson, E.~Agustsson, R.~Timofte, and L.~Van~Gool.
\newblock Generative adversarial style transfer networks for face aging.
\newblock In {\em Proceedings of the IEEE Conference on Computer Vision and
  Pattern Recognition Workshops}, pp. 2084--2092, 2018.

\bibitem{pennington2014glove}
J.~Pennington, R.~Socher, and C.~Manning.
\newblock Glove: Global vectors for word representation.
\newblock In {\em Proceedings of the 2014 conference on empirical methods in
  natural language processing (EMNLP)}, pp. 1532--1543, 2014.

\bibitem{peters2018deep}
M.~E. Peters, M.~Neumann, M.~Iyyer, M.~Gardner, C.~Clark, K.~Lee, and
  L.~Zettlemoyer.
\newblock Deep contextualized word representations.
\newblock {\em arXiv preprint arXiv:1802.05365}, 2018.

\bibitem{saito2016pinscreen}
S.~Saito, L.~Wei, J.~Fursund, L.~Hu, C.~Yang, R.~Yu, K.~Olszewski, S.~Chen,
  I.~Benavente, Y.-C. Chen, et~al.
\newblock Pinscreen: 3d avatar from a single image.
\newblock In {\em SIGGRAPH ASIA 2016 Emerging Technologies}, p.~15. ACM, 2016.

\bibitem{saito2017photorealistic}
S.~Saito, L.~Wei, L.~Hu, K.~Nagano, and H.~Li.
\newblock Photorealistic facial texture inference using deep neural networks.
\newblock In {\em Proceedings of the IEEE Conference on Computer Vision and
  Pattern Recognition}, pp. 5144--5153, 2017.

\bibitem{sutskever2014sequence}
I.~Sutskever, O.~Vinyals, and Q.~V. Le.
\newblock Sequence to sequence learning with neural networks.
\newblock In {\em Advances in neural information processing systems}, pp.
  3104--3112, 2014.

\bibitem{thies2016face2face}
J.~Thies, M.~Zollhofer, M.~Stamminger, C.~Theobalt, and M.~Nie{\ss}ner.
\newblock Face2face: Real-time face capture and reenactment of rgb videos.
\newblock In {\em Proceedings of the IEEE Conference on Computer Vision and
  Pattern Recognition}, pp. 2387--2395, 2016.

\bibitem{vaswani2017attention}
A.~Vaswani, N.~Shazeer, N.~Parmar, J.~Uszkoreit, L.~Jones, A.~N. Gomez,
  {\L}.~Kaiser, and I.~Polosukhin.
\newblock Attention is all you need.
\newblock In {\em Advances in neural information processing systems}, pp.
  5998--6008, 2017.

\bibitem{wang2018video-to-video}
T.~Wang, M.~Liu, J.~Zhu, G.~Guilin, A.~J. Tao, J.~Kautz, and B.~Catanzaro.
\newblock Video-to-video synthesis.
\newblock {\em neural information processing systems}, pp. 1144--1156, 2018.

\bibitem{wang2018high-resolution}
T.~Wang, M.~Liu, J.~Zhu, A.~J. Tao, J.~Kautz, and B.~Catanzaro.
\newblock High-resolution image synthesis and semantic manipulation with
  conditional gans.
\newblock {\em computer vision and pattern recognition}, pp. 8798--8807, 2018.

\bibitem{wang2018pix2pixHD}
T.~C. Wang, M.~Y. Liu, J.~Y. Zhu, A.~Tao, and B.~Catanzaro.
\newblock High-resolution image synthesis and semantic manipulation with
  conditional gans.
\newblock 2017.

\bibitem{wu2018reenactgan}
W.~Wu, Y.~Zhang, C.~Li, C.~Qian, and C.~C. Loy.
\newblock Reenactgan: Learning to reenact faces via boundary transfer.
\newblock {\em european conference on computer vision}, pp. 622--638, 2018.

\bibitem{xu2017face}
R.~Xu, Z.~Zhou, W.~Zhang, and Y.~Yu.
\newblock Face transfer with generative adversarial network.
\newblock {\em arXiv preprint arXiv:1710.06090}, 2017.

\bibitem{zhou2015end}
J.~Zhou and W.~Xu.
\newblock End-to-end learning of semantic role labeling using recurrent neural
  networks.
\newblock In {\em Proceedings of the 53rd Annual Meeting of the Association for
  Computational Linguistics and the 7th International Joint Conference on
  Natural Language Processing (Volume 1: Long Papers)}, vol.~1, pp. 1127--1137,
  2015.

\bibitem{zhu2017unpaired}
J.~Zhu, T.~Park, P.~Isola, and A.~A. Efros.
\newblock Unpaired image-to-image translation using cycle-consistent
  adversarial networks.
\newblock {\em international conference on computer vision}, pp. 2242--2251,
  2017.

\end{thebibliography}
\end{document}